\documentclass{anstrans}
\title{A Methodology for Thermal Limit Bias Predictability through Artificial Intelligence}
\author{Anirudh Tunga,$^{*}$ M. J. Mueterthies,$^{*,\dagger}$, and J.Nistor,$^{*,\dagger}$} 

\institute{
$^{*}$Blue Wave AI Labs, 1284 Win Hentschel Blvd, West Lafayette, IN 47906, anirudh.tunga@bwailabs.com 
\and
$^{\dagger}$Department of Physics \& Astronomy, 525 Northwestern Avenue, Purdue University, West Lafayette, IN 47907
\and
}

\usepackage{graphicx} 
\usepackage{booktabs} 
\usepackage{microtype} 
\usepackage{array}

\newcommand{\R}{\mathbb{R}}

\begin{document}
\section{Introduction}
Operational limits on nuclear power plants are set to ensure the plant can be safely operated without posing any unnecessary risks to public health and safety. Thermal limits are established to maintain the integrity of the fuel cladding in both normal and transient conditions \cite{GE_Systems_Manual}. Three thermal limits are established for a Boiling Water Reactor (BWR). Two of the limits - Linear Heat Generation (LHGR) limit and Critical Power Ratio (CPR) limit – ensure the integrity of the fuel cladding under normal operations and transient events by limiting fuel rod power density and by maintaining nucleate boiling around the fuel rods. The third limit –  Average Planar Linear Heat Generation Rate (APLHGR) limit – maintains the core geometry during postulated accidents by minimizing the gross fuel cladding failure \cite{GE_Systems_Manual}. 

In a nuclear power plant (NPP), a core monitoring system is used to calculate and monitor significant plant parameters including thermal limits. These parameters are generally available in two modes -- offline and online. In the offline mode, the core simulator has no access to information from the in-core nuclear instrumentation. During the core design and planning phase of a fuel cycle, all the parameters are necessarily calculated in an offline mode. In contrast, in an online mode, the core monitoring system leverages the in-core neutron flux measurements to calculate nodal fit coefficients for the online power distribution. These coefficients are used to perform an adaptive process on the offline parameters to arrive at online parameters, which reflect the operating conditions more accurately. As online parameters can only be determined for a cycle that is in operation, reactor core design can only be accomplished with access to offline parameters determined from the core simulator. 

\begin{figure}[ht!]
  \centering
  \includegraphics[scale=0.54]{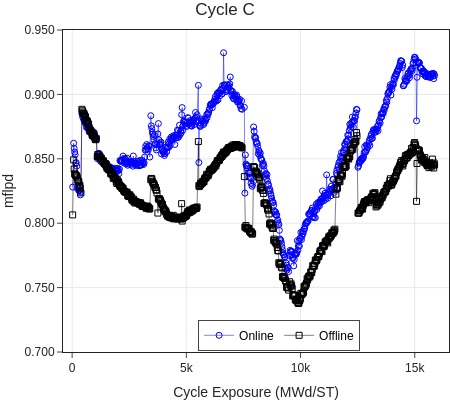}
  \caption{\small Online (blue circles) and offline (black squares) max $MFLPD$ for one complete fuel cycle at a commercially operating BWR. The gap between the two is referred to as the thermal limit bias.}
  \label{fig:bias}
\end{figure}

Historically, the offline and online thermal limits have exhibited large and inconsistent deviations from one another, as can be seen in Fig. \ref{fig:bias}. This deviation is often referred to as thermal limit bias. The thermal limit bias is not constant and occurs at various magnitudes throughout the fuel cycle. Moreover, the extent of the bias varies unpredictably from one fuel cycle to the next, resulting in a challenge to adequately account for this bias during reactor core design. Currently, there is no method to accurately predict this bias. To account for the uncertainty in the bias, NPPs have significantly reduced their design goals to prevent online thermal limits from exceeding the administrative thermal limits. This conservative design margin leads to fuel cycle efficiency loss if the online thermal limits are significantly below the administrative limits and lead to operational challenges if the added margin is insufficient during reactor operation. The high thermal limit bias between offline and online methods leads to many operational challenges, including (i) increased fuel costs due to larger reload batch sizes or increased enrichment, (ii) unplanned use of shallow control rods leading to loss of full power energy capability, (iii) unplanned rod pattern changes resulting in reactivity management challenges, and (iv) unplanned power de-rates. 

Having the ability to accurately predict the thermal limit bias (or more accurately predict the true online thermal limits) during the design phase will allow NPPs to reduce the thermal limit margin at various parts of the cycle while still satisfying the administrative limits and optimizing the fuel cycle economics. To address this challenge, we propose a deep learning-based model, which takes the offline parameters and offline thermal limits as inputs, corrects the bias and predicts new thermal limits that are closer to the online thermal limits. We utilize historical offline and online data from a currently operating BWR to train our model. Our model architecture is based on a fully convolutional encoder-decoder neural network. 

In this paper, the scope is limited to the LHGR limit. The core monitoring system uses the Maximum Fraction of Limiting Power Density ($\mathrm{MFLPD}$) \cite{GE_Systems_Manual}, as defined in Eq. \ref{eq:lhgr}, to track the LHGR limit. In Eq. \ref{eq:lhgr}, MRPD is the Maximum Rod Power Density and RPDLM is the Thermal Mechanical Limit. 
\begin{equation}\label{eq:lhgr}
    \mathrm{MFLPD} = \frac{\mathrm{MRPD}}{\mathrm{RPDLM}} \frac{\mathrm{Maximum \; LHGR_{actutal}}}{\mathrm{LHGR \; Limit}}
\end{equation} 

\begin{figure*}[ht!]
  \centering
  \includegraphics[scale=0.44]{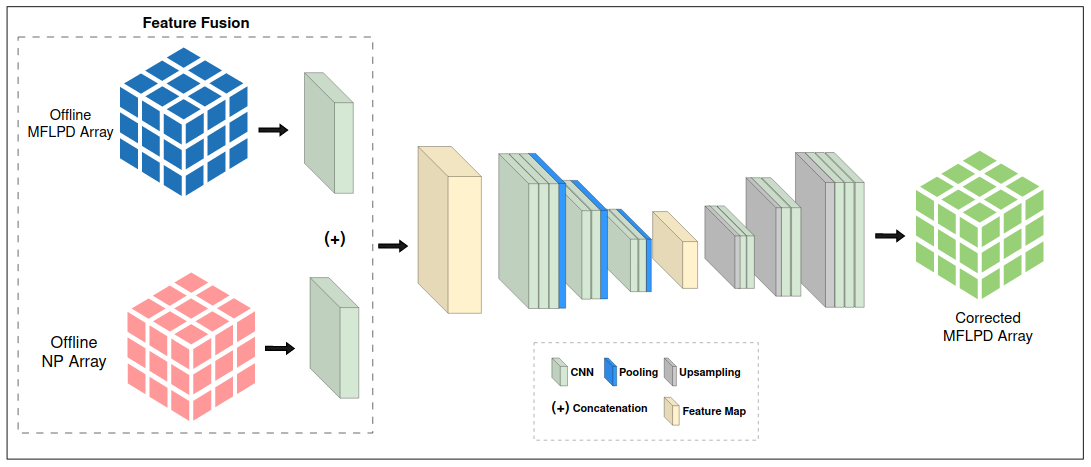}
  \caption{Overview of the proposed model architecture. Feature fusion is performed on the input features before passing them through the encoder and the decoder. }
  \label{fig:arch}
\end{figure*}

We evaluate our model on the five most recent fuel cycles. Our model, on average, reduces the bias between limiting values of offline and online $\mathrm{MFLPD}$ by $72\%$. A variant of our proposed model has been commercially deployed for the management of future fuel cycles.

\section{Methodology}

\subsection{Dataset}

The dataset used in this work is from a currently operating large BWR-4 (with Mark I containment). The dataset includes online nodal MFLPD values (MFLPD$_{online} \in \R ^{H \times W \times D}$), offline nodal MFLPD values (MFLPD$_{offline} \in \R ^{H \times W \times D}$), offline Nodal Power values ($NP_{offline} \in \R ^{H \times W \times D}$), and other parameters computed by the commercial core simulator. The dataset contains the data for 11 fuel cycles. The model uses MFLPD$_{offline}$ and a combination of other offline nodal parameters as inputs and MFLPD$_{online}$ as target.

Although nodal MFLPD values are important, NPPs track the limiting value of the nodal MFLPD arrays more closely---online MFLPD limit ($mflpd_{online} \in \R$) and offline MFLPD limit ($mflpd_{offline} \in \R$) as defined in Eq. \ref{eq:mflpd}.

\begin{equation}\label{eq:mflpd}
\begin{split}
    mflpd_{online} = \max (MFLPD_{online} ) \\
    mflpd_{offline} = \max (MFLPD_{offline})
\end{split}
\end{equation} 

\subsection{Proposed Architecture}

Our proposed model is a fully convolutional neural network, which consists of a feature fusion network, an encoder network, and a corresponding decoder network. Our encoder and decoder networks are based on the SegNet \cite{badrinarayanan2017segnet} architecture, which has shown success in learning spatial relationships within the data and learning from low-resolution feature maps. The feature fusion network, encoder, and decoder are jointly trained in an end-to-end manner.

The feature fusion network consists of two separate blocks of convolutional layers - one for MFLPD$_{offline}$, and the other for $NP_{offline}$. Individual feature maps are extracted from the inputs, and the feature maps are concatenated in the feature fusion network. The concatenated feature map is the input for the encoder-decoder network.

The encoder network consists of 3 stages and a total of 7 convolutional layers. At each stage, convolutional layers are applied to the input feature map to produce a new feature map. The new feature maps are batch normalized \cite{ioffe2015batch} and a non-linear activation ReLu is applied to them. At each stage, max-pooling is applied such that the spatial dimensions of the feature maps are reduced by a factor of 2. At each max-pooling step, the location of the max values of the features is stored in memory, which is later used by the decoder layers to reconstruct the output.

For each of the 7 encoder layers, there is a corresponding decoder layer. Following the encoder network, the decoder network also has 3 stages. The decoder, at each stage, up-samples the input feature map by utilizing the stored location of the max values of the features, which were stored during the encoding step. The sparse feature map produced by up-sampling is then passed through decoder convolutional layers to produce dense feature maps. The dense feature maps are then batch normalized. The final high-dimensional feature map is only passed through a convolutional layer with no batch normalization or non-linear activation, to output a corrected nodal MFLPD array (MFLPD$_{predicted}$), which is of the same shape as the MFLPD$_{online}$ array.

\begin{table*}[!ht]
  \centering
  \caption{\small Quantitative comparison of our model with the offline methods. We quantify the performance using three metrics - i) mean squared error between nodal arrays, ii) mean absolute difference between limiting values, and iii) maximum absolute difference between the limiting values.}
  \label{table:tl_res} 
  \begin{tabular}{|c | >{\centering}p{0.1\textwidth} >{\centering}p{0.1\textwidth} | >{\centering}p{0.1\textwidth} >{\centering}p{0.1\textwidth} | >{\centering}p{0.1\textwidth} c|} \hline 
  \multicolumn{1}{| c |}{{Cycles}} & \multicolumn{2}{c|}{{Nodal $MFLPD$}} & \multicolumn{2}{c|}{{$mflpd$ limit}}  & \multicolumn{2}{c|}{{Max $mflpd$ bias}}\\\hline
   & \multicolumn{2}{c|}{{$MSE \; (10^{-4})$}} & \multicolumn{2}{c|}{{$MAE \; (10^{-2})$}} & \multicolumn{2}{c|}{{$Absolute \; difference \; (10^{-2})$}} \\\hline
    & {Offline} & {Model} & {Offline} & {Model}  & {Offline} & {Model}  \\\hline
    \  Cycle A  &  $4.18$ & $\pmb{1.65}$ &  $3.44$ & $\pmb{0.95}$ & $7.37$ & $\pmb{3.63}$       \\ \hline
    \ Cycle B & $6.10$ &  $\pmb{1.14}$ & $2.88$ & $\pmb{0.99}$ & $7.39$ & $\pmb{3.72}$\\ \hline 
    \ Cycle C & $7.74$ &  $\pmb{1.41}$ & $3.93$ & $\pmb{1.03}$ & $8.81$ & $\pmb{4.48}$\\ \hline 
    \ Cycle D & $8.42$ &  $\pmb{2.28}$ & $5.14$ & $\pmb{1.61}$ & $9.17$ & $\pmb{3.41}$\\ \hline 
    \ Cycle E & $3.80$ &  $\pmb{0.89}$ & $4.45$ & $\pmb{0.81}$ & $8.05$ & $\pmb{4.01}$\\ \hline 
  \end{tabular}
\end{table*}

\subsection{Training}
The proposed model has been implemented in PyTorch \cite{paszke2019pytorch}. We trained the model using AdamW \cite{loshchilov2017decoupled} optimizer with a weight decay of $0.01$, and a 1-cycle learning rate policy \cite{smith2019super} is used with $max\_lr = 0.005$. We use the mean squared error loss as the loss function for training the model. We use a boolean reactor mask to select only the valid elements of the three-dimensional nodal arrays. We need to use a mask as the data from the octagon-shaped reactor core has padding when represented as a three-dimensional array. The reactor mask is applied to both $MFLPD_{online}$ and $MFLPD_{predicted}$, before calculating the mean squared error loss. For the inputs to the model, there were multiple options available in addition to the $MFLPD_{offline}$. We performed ablation experiments using only $MFLPD_{offline}$, and adding other inputs. Adding $NP_{offline}$ performed the best among all the experiments so it was chosen in the final model architecture. All the experiments were run on a single NVIDIA A10 GPU.

\section{Results and Analysis}

\subsection{Experiments}
To evaluate our proposed model, we performed five different experiments by selecting one of the five latest fuel cycles as an independent test cycle in each experiment. In each of the experiments, the test cycle is separated from the dataset and the rest of the data is used for training and validation. We used $70\%$ of the data for training and the remaining $30\%$ for validation. The training set is shuffled before each epoch. We select and save the best-performing model on the validation set as the final model. The final model is evaluated on the test model. The training and testing processes are independent. 

\subsection{Results}
We use three metrics to quantitatively compare our model with the offline methods, and evaluate the model's ability to correct the offline values. The online $MFLPD$ values are the targets, so the model and the offline methods are evaluated for their closeness to the online values. The three metrics are: (i) the mean squared error between the three-dimensional nodal arrays, (ii) the mean absolute difference between the limiting values ($mflpd$), and (iii) the maximum absolute difference between the limiting values (max bias) in a fuel cycle. In Table \ref{table:tl_res} we report these three metrics for both offline methods and our model on each of the five test cycles. The first metric - mean squared error between nodal arrays - provides a measure of closeness between the two nodal arrays. Since this metric is calculated for the entire three-dimensional arrays, it evaluates the two arrays at all locations of the reactor core. In Table \ref{table:tl_res}, we can see that the mean squared error between $MFLPD_{online}$ and $MFLPD_{model}$ is significantly reduced as compared to the error between $MFLPD_{online}$ and $MFLPD_{offline}$. On average, our proposed model reduces the error between nodal arrays by $74\%$.

\begin{figure*}[ht!]
  \centering
  \includegraphics[scale=0.45]{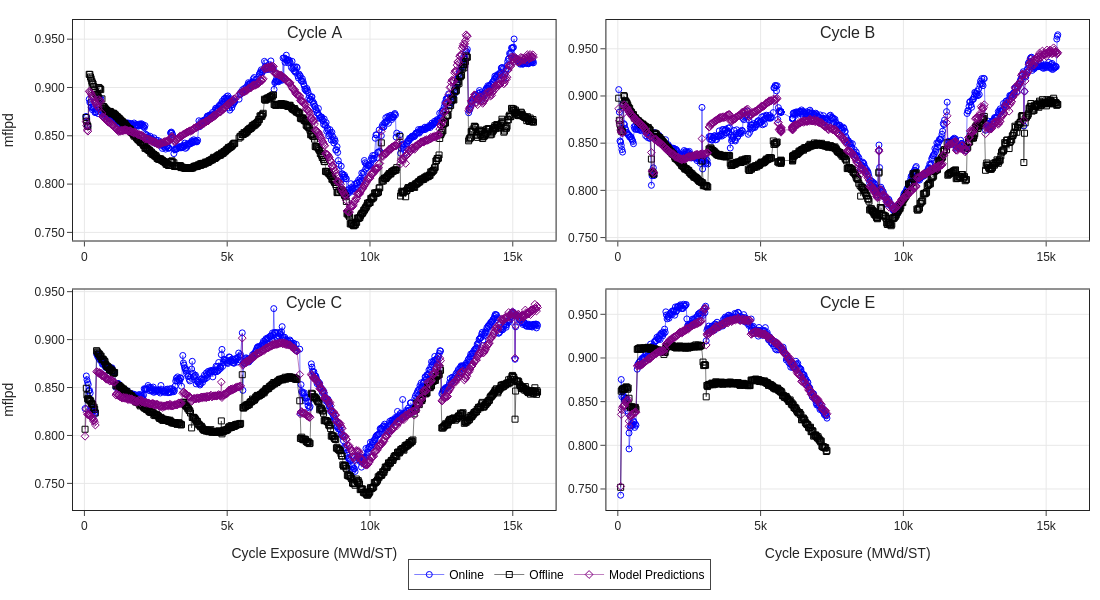}
  \caption{Results on the four recent independent test cycles. Cycle E was still in progress when experiments were performed. The model corrects the bias in all the test cycles, throughout the cycle.}
  \label{fig:res}
\end{figure*}

The second metric, the mean absolute difference between the limiting values, is the most important metric for operating a NPP. The deviation between online limits ($mflpd_{online}$) and offline limits ($mflpd_{offline}$) can lead to operational challenges, such as unplanned power de-rated conditions, use of unplanned control blades, or increased fuel costs by loading more fuel than required for targeted energy production. In Fig. \ref{fig:res}, we can see the deviation between $mflpd_{online}$ and $mflpd_{offline}$ throughout the fuel cycle, in four of the test cycles. We can also see the ability of the model to reduce the deviation consistently throughout the cycle. In Fig. \ref{fig:res}, only a partial Cycle E is shown as the fuel cycle was still in operation when the experiments were performed. In Table \ref{table:tl_res}, we have quantified the deviation between $mflpd_{online}$ and $mflpd_{offline}$, and the deviation between $mflpd_{online}$ and $mflpd_{predicted}$. We see that our proposed model has reduced this deviation in all of the five test cycles. As compared to the offline methods, the deviation is reduced by $72\%$ on average by our model. The third metric is an extension of the second metric, where the focus is on the maximum absolute difference between limiting values in the cycle instead of the mean error over the cycle. This metric tracks the worst performance in a fuel cycle, and a large value of this metric causes a lot of uncertainty in the operations of the NPP. The model reduces this maximum deviation by $52\%$ on average.

The results on all three metrics show that the proposed model is successfully able to predict and correct the bias between the offline and online $MFLPD$ for all the test cycles. For a model to be successful in practical use cases like cycle management, the model needs to perform well on future fuel cycles. This was our motivation to evaluate the model on the recent independent fuel cycles, which presents a greater challenge. This evaluation method was preferred over evaluating on an in-distribution test set, for example, a test set derived by splitting the dataset as $70\%$ training, $20\%$ validation, and $10\%$ test sets. The successful and robust results of the proposed model have enabled a variant of the model to be deployed commercially for fuel cycle management in BWRs.


\section{Conclusion}

In this work, we present a fully convolutional encode-decoder model, which corrects the thermal limit bias between the offline and online $MFLPD$. The proposed model helps the NPP operators to plan more economical fuel cycles, by greatly reducing the magnitude of thermal limit bias. We evaluated the model on five recent test cycles using three different metrics, each metric measuring a different aspect of the bias in a fuel cycle. We demonstrate that the model performs exceptionally well on all test cycles and all the metrics as compared to offline methods. We have also commercially deployed a variant of this model to a currently operating BWR. 

In the future, we would like to extend this model to other thermal limits - CPR and APLHGR. We also plan to evaluate the model's performance when we combine data from multiple BWRs, and in cases where data for only a few historical cycles is available.

\appendix

\section{Acknowledgments}
The authors would like to acknowledge the U.S. Department of Energy for supporting many industry-led efforts through various funding pathways established within the {\em U.S. Industry Opportunities for Advanced Nuclear Technology Development} (DE-FOA-0001817) program. This material is based upon work supported by the Department of Energy under award DE-NE0008930. Neither the United States Government nor any agency thereof, nor any of their employees, makes any warranty, express or implied, or assumes any legal liability or responsibility for the accuracy, completeness, or usefulness of any information, apparatus, product, or process disclosed, or represents that its use would not infringe privately owned rights. The views and opinions of authors expressed herein do not necessarily state or reflect those of the United States Government or any agency thereof.

\bibliographystyle{ans}\small
\bibliography{bibliography}

@article{badrinarayanan2017segnet,
  title={SegNet: A deep convolutional encoder-decoder architecture for image segmentation},
  author={Badrinarayanan, Vijay and Kendall, Alex and Cipolla, Roberto},
  journal={IEEE transactions on pattern analysis and machine intelligence},
  volume={39},
  number={12},
  pages={2481--2495},
  year={2017},
  publisher={IEEE}
}

@inproceedings{ioffe2015batch,
  title={Batch normalization: Accelerating deep network training by reducing internal covariate shift},
  author={Ioffe, Sergey and Szegedy, Christian},
  booktitle={International conference on machine learning},
  pages={448--456},
  year={2015},
  organization={pmlr}
}

@article{paszke2019pytorch,
  title={PyTorch: An imperative style, high-performance deep learning library},
  author={Paszke, Adam and Gross, Sam and Massa, Francisco and Lerer, Adam and Bradbury, James and Chanan, Gregory and Killeen, Trevor and Lin, Zeming and Gimelshein, Natalia and Antiga, Luca and others},
  journal={Advances in neural information processing systems},
  volume={32},
  year={2019}
}

@article{loshchilov2017decoupled,
  title={Decoupled weight decay regularization},
  author={Loshchilov, Ilya and Hutter, Frank},
  journal={arXiv preprint arXiv:1711.05101},
  year={2017}
}

@techreport{GE_Systems_Manual,
  title = {General Electric Systems Technology Manual},
  chapter = {1.8 Thermal Limits},
  institution = {U.S. Nuclear Regulatory Commission},
  url = {https://www.nrc.gov/docs/ML1125/ML11258A297.pdf},
  note = {Accessed: 26 June 2024}
}

@inproceedings{smith2019super,
  title={Super-convergence: Very fast training of neural networks using large learning rates},
  author={Smith, Leslie N and Topin, Nicholay},
  booktitle={Artificial intelligence and machine learning for multi-domain operations applications},
  volume={11006},
  pages={369--386},
  year={2019},
  organization={SPIE}
}
\end{document}